\documentclass[runningheads]{llncs}

\usepackage[T1]{fontenc}
\usepackage{graphicx,verbatim}
\usepackage{subcaption}
\usepackage{float}
\usepackage{amsmath}
\usepackage{amssymb}
\usepackage{xcolor}
\usepackage{url}

\begin{document}

\title{Implicit Neural Representations for Multimodal Longitudinal Image Imputation and Interpolation}
\titlerunning{INRs for multimodal longitudinal image imputation}

\author{Sina Wendrich\inst{1,2},
    Lukas Förner\inst{1,2,3,4},
    Zoe Reinke\inst{1,2},
    Kartikay Tehlan\inst{1,2,3,4},
    Ansgar Berlis\inst{1},
    Michael Frühwald\inst{5},
    Matthias Wagner\inst{1,6},
    Thomas Wendler\inst{1,2,3,4,6}}

\authorrunning{S. Wendrich et al.}

\institute{
Department of Diagnostic and Interventional Radiology and Neuroradiology, University Hospital Augsburg, Augsburg, Germany \and
Digital Medicine, University Hospital Augsburg, Augsburg, Germany \and
Chair for Computer-Aided Medical Procedures and Augmented Reality, Technical University of Munich, Garching bei München, Germany \and
Bavarian Cancer Research Center (BZKF), Augsburg, Germany \and
Pediatrics and Adolescent Medicine, Swabian Children's Cancer Center, Augsburg, Germany \and
Center of Advanced Analytics and Predictive Sciences, University of Augsburg, Augsburg, Germany
}

\maketitle
\begin{abstract}
Longitudinal multiparametric MRI is central to follow-up imaging in oncology, yet real-world clinical data are characterised by missing sequences, heterogeneous acquisition protocols, and varying spatial resolutions across time points. We propose a patient-specific conditional implicit neural representation (INR) that models multimodal longitudinal MRI as a continuous function of world coordinates, time, and modality conditioning. The model is trained with stochastic modality dropout to handle incomplete data, and its continuous coordinate-space formulation enables both spatial and temporal interpolation without resampling to a fixed voxel grid. A self-consistency-based confidence estimator is derived from cross-modal reconstruction performance at inference time. We evaluate the framework on longitudinal MRI from paediatric brain tumour patients, demonstrating statistically significant improvements over linear interpolation for T1CE and FLAIR ($p < 0.05$), with mean MS-SSIM of $0.95 \pm 0.02$ for T1CE. Predicted confidence correlates strongly with true reconstruction quality (Pearson $r$ up to $0.996$), suggesting reliable deployment potential in heterogeneous clinical settings.
Implementation available at: \url{https://github.com/SinaWend/ATRT_INR_Modality_Imputation}.
\end{abstract}


\section{Introduction}

There are multiple clinical indications in oncology where closely following the development of recurrence after primary therapies such as surgery, radiotherapy, or first-line systemic therapy requires regular follow-up imaging. In particular, in the case of brain tumours, this is typically performed using multiparametric MRI~\cite{Villanueva-Meyer}. However, due to cost constraints or differences in institutional imaging protocols, not all sequences are available or acquired at every time point. Furthermore, data are often acquired across different scanners and institutions with varying imaging resolutions and acquisition parameters, making voxel-wise comparison across time points non-trivial even after image registration. Patient positioning is also not kept identical across examinations, such that image registration is required for voxel-wise comparison.
Voxel-wise analysis of temporal changes, however, could provide highly valuable information for outcome prediction, therapy planning, and prognosis.

In this paper, we propose leveraging longitudinal multiparametric MRI data acquired in real-world clinical settings, where particular sequences may be missing at different time points, to train a multiparametric four-dimensional implicit neural representation (INR). Since INRs operate in a continuous coordinate space, the proposed model can not only impute missing sequences across time and modalities, but also perform spatial interpolation. The latter is done synthesizing images at any resolution or voxel spacing, and temporal interpolation, which enables estimating image appearance at time points between actual acquisitions. The former directly addresses the challenge of varying acquisition resolutions across scanners and institutions, while the latter enables alignment of asynchronous longitudinal series. Such a representation could serve as an upstream component for image-based disease progression prediction, therapy planning, or related downstream clinical tasks.

The contributions of this paper are as follows: (i) a continuous coordinate-based implicit neural representation for multimodal longitudinal MRI, enabling imputation of missing sequences as well as spatial and temporal interpolation; (ii) a training strategy and loss formulation that accommodate incomplete longitudinal data, where not all sequences are available at each time point; (iii) methods for estimating the confidence of the imputed and interpolated values at any given spatial and temporal location; and (iv) an evaluation of the proposed methodology on real-world cases of paediatric brain tumours.

\section{Related Work}

\paragraph{Patient-Specific INRs for Longitudinal MRI.}
INRs can be fitted directly to a single subject without large-scale population training. NeRP exploits this by embedding a longitudinal prior image, a previously acquired scan of the same patient, into the network weights as initialisation, then optimising the INR to reconstruct a subsequent sparsely sampled scan~\cite{Shen}. This demonstrates that INRs can incorporate patient-specific longitudinal information. However, NeRP operates on a single imaging sequence and focuses on sparse reconstruction from raw measurements, rather than modelling the patient's imaging trajectory across multiple sequences and time points.

\paragraph{Spatiotemporal Modelling.} 
Spatiotemporal INRs have shown particular promise for dynamic MRI reconstruction, enabling unsupervised learning from sparsely sampled k-space data by modelling image intensity as a continuous function over both space and time~\cite{Feng}. The use of INRs for Cine MRI reconstruction has similarly been explored, combining spatial and temporal implicit representations for motion-resolved cardiac imaging~\cite{Shao}. However, in both cases the temporal axis operates on short time-resolution scales within a single scan to combat motion artefacts, leaving a gap in longitudinal modelling across clinical time points spanning weeks to months.

\paragraph{Cross-Sequence Synthesis and Translation.}
Missing modality synthesis has been widely studied using convolutional and adversarial architectures, ranging from modality-invariant encoder-decoders~\cite{Chartsias} to advanced fusion and synthesis frameworks~\cite{Zhou,Liu,Dorent}. However, these approaches operate on fixed voxel grids and typically assume consistent spatial resolution and preprocessing across subjects. Coordinate-based neural fields offer an alternative by representing multimodal MRI directly in world coordinates as continuous functions, enabling joint modelling across sequences~\cite{Chen}. While this demonstrates the feasibility of INRs for multi-sequence settings and removes the dependency on fixed grids, it remains restricted to a single time point per patient. The combination of multi-sequence imputation and longitudinal interpolation within a patient-specific coordinate-based framework remains unexplored.

\paragraph{Confidence Estimation.}
Reliable deployment of deep learning models in medicine requires principled uncertainty estimation. Classical approaches include Bayesian neural networks and Monte Carlo dropout~\cite{Gal}, and deep ensembles~\cite{Lakshminarayanan}, which estimate epistemic uncertainty via model variability. These methods generally require multiple stochastic forward passes or ensemble training.
An alternative derives confidence from reconstruction consistency, where higher reconstruction error indicates lower model certainty. Our work follows this principle by computing a confidence score from cross-modal self-reconstruction performance for available sequences as proxy for the confidence of the imputed sequence, without requiring Bayesian modelling or multiple model instantiations.

\section{Methods}

\subsection{Data}

We consider longitudinal imaging data from 5 paediatric patients with brain tumours following surgery. The data originate from a region-wide multi-centre register for which we have ethical approval to analyse. For each patient, on average nine multiple post-operative time points are available, at which volumetric MRI scans were acquired as part of routine clinical follow-up.

Acquired sequences typically include some of the following MRI sequences: T1-weighted (T1w), T2-weighted (T2w), contrast-enhanced T1-weighted (T1w-CE), T2-weighted fluid-attenuated inversion recovery (FLAIR), and diffusion-weighted image (DWI)-derived apparent diffusion coefficient (ADC) maps. However, not all sequences are consistently available at every time point, reflecting real-world clinical acquisition variability. Moreover, scans were acquired on different scanners with heterogeneous acquisition parameters and spatial resolutions.

\subsection{Pre-processing and Registration}
\label{sec:prep}
All images are spatially aligned across time points using rigid registration.  The first available time point after resection is used as the reference for each patient, and its brain mask serves as the fixed image for registration. Brain segmentation is performed using HD-BET~\cite{hd-bet} when no mask is available, followed by connected component analysis to retain only the largest brain component.

Rigid registrations are computed across time points based on the binary brain masks using ITK-Elastix~\cite{elastix}. The estimated rigid transformation matrices are subsequently applied to all corresponding sequences of the respective time 
point, thereby bringing all images into a common coordinate system.

Importantly, no resampling of the image volumes is performed. Instead, the rigid transformation is applied by updating the image header information (origin and direction), while retaining the original voxel grids and spatial resolution. This preserves the native acquisition characteristics of each sequence while enabling the INR to learn directly from heterogeneous coordinate spaces.

Within each individual time point, all sequences are assumed to be co-registered, as confirmed by visual inspection for every patient in the cohort used in this work. After spatial alignment, all volumes are intensity-normalised on a per-sequence basis to the range [0, 1] across all time points of a patient.

\subsection{Conditional Implicit Neural Representation}

For each patient, we learn a continuous function that maps spatial location, time, and available modality information to multimodal intensities. The model operates directly in world coordinates and does not require a fixed voxel grid.

\paragraph{Input conditioning vector}

Let $\mathbf{x} \in \mathbb{R}^3$ denote the world coordinate (in millimetres) of a voxel and let $t \in [0,1]$ denote the normalised time coordinate with a known mapping to days since surgery. 

At a given spatial location and time point, some modalities may be available while others are missing. Let $\mathcal{M} = \{\text{T1}, \text{T1CE}, \text{T2}, \text{FLAIR}, \text{ADC}\}$ denote the full modality set with $|\mathcal{M}| = 5$. For each modality $m \in \mathcal{M}$ we define:

\[
c_m =
\begin{cases}
I_m(\mathbf{x}, t) & \text{available for training,} \\
0 & \text{otherwise,}
\end{cases}
\qquad
b_m =
\begin{cases}
1 & \text{used in training,} \\
0 & \text{otherwise.}
\end{cases}
\]

The conditioning vector is then defined as $c(x,t) = [c_1, \dots, c_5, b_1, \dots, b_5] \in \mathbb{R}^{10}$, and 
encodes which modalities are available at a given time point during training. Since zero is a valid normalised intensity, the binary flags $b_m$ are essential to disambiguate from a genuinely zero-intensity observation.

\paragraph{Spatial Encoding.}

Spatial coordinates are first transformed into multiple levels using a multi-resolution hash grid encoding. 
At each level, the continuous coordinate is scaled to the corresponding grid resolution. Features are retrieved from the hashed embedding table and combined using trilinear interpolation. This produces a compact representation capable of modelling fine anatomical detail.

The three-dimensional coordinate $\mathbf{x}$ is mapped to sixteen resolution levels. Each level stores four learnable feature channels. This yields a spatial embedding vector $\psi(x) \in \mathbb{R}^{64}$ that is used as the spatial input for the INR.

\paragraph{Temporal Encoding.}

The scalar temporal coordinate $t$ is encoded using Fourier positional encoding with 8 frequency bands:
\[
\phi(t) = 
\left[
t,
\left(\sin(2^k \pi t),
\cos(2^k \pi t)
\right)_{k=0}^{7}
\right]
\in \mathbb{R}^{17}.
\]

This encoding enables the network to resolve fine-grained temporal variations that a raw scalar coordinate cannot.

\paragraph{Full Network Architecture.}

The network consists of a fully connected ReLU multi-layer perceptron with 5 hidden layers of width 256. A mid-layer skip connection concatenates the original input embedding with intermediate activations, improving gradient flow and representation capacity. It learns a mapping
\[
f_\theta : \mathbb{R}^{91} \to \mathbb{R}^{5}, \quad (\psi(x),\, \phi(t),\, c(x,t)) \mapsto \hat{\mathbf{I}}(\mathbf{x}, t),
\]
where the output vector $\hat{\mathbf{I}} = (\hat{I}_{\text{T1}},\, \hat{I}_{\text{T1CE}},\, \hat{I}_{\text{T2}},\, \hat{I}_{\text{FLAIR}},\, \hat{I}_{\text{ADC}})$ contains the predicted intensities for all five modalities simultaneously. The 91 dimensions result from 64 spatial, 17 time and 5 MRI sequence encodings, and 5 binary flags.

\subsection{Training Strategy}

Let $N$ denote the number of patients in the cohort. For each patient, a dedicated implicit neural representation is trained independently using all available longitudinal data. Let $\mathcal{T}$ denote the set of available time points for a given patient, and let $\mathcal{M}_t \subseteq \mathcal{M}$ denote the set of modalities available at time point $t \in \mathcal{T}$. During training, time points are sampled uniformly from $\mathcal{T}$.

To enable robust reconstruction of missing modalities, we employ a stochastic modality dropout strategy. At each optimisation step and for a sampled time point $t$, a non-empty strict subset $\mathcal{D}_t \subset \mathcal{M}_t$ is selected uniformly at random for time points with $\mathcal{M}_t\ge2$. The modalities in $\mathcal{D}_t$ are removed from the conditioning input, while the remaining modalities $\mathcal{A}_t = \mathcal{M}_t \setminus \mathcal{D}_t$ are provided to the network via the conditioning vector.

For spatial locations $\mathbf{x}$ sampled from the corresponding registered volumes, the network predicts intensities for all modalities. The loss is computed exclusively for the dropped modalities and averaged over the batch size:
\[
\mathcal{L}_{1} =
\frac{1}{|\mathcal{D}_t|}
\sum_{m \in \mathcal{D}_t}
\left|
\hat{I}_m(\mathbf{x}, t) - I_m(\mathbf{x}, t)
\right|.
\]

This procedure ensures that the network is repeatedly exposed to incomplete modality configurations and is forced to infer missing information from the remaining sequences. In contrast to deterministic masking, stochastic modality dropout generates a diverse set of conditioning combinations during training, thereby promoting cross-modal consistency and preventing identity mappings.

\subsection{Evaluation and Confidence Estimation}

Per patient, one interior time point $t^\ast$ with all modalities available is selected and held out. Let $\mathcal{H} = \{\text{ADC}, \text{FLAIR}, \text{T1CE}\}$ denote the set of candidate held-out modalities. T1w and T2w are assumed consistently available. 
The hold-out time lies strictly between the first and last examinations, restricting evaluation to temporal interpolation. The respective modality is excluded from the conditioning input at $t^\ast$ during training to avoid data leaks. At inference, let $h \in \mathcal{H}$ denote the held-out modality and $\mathcal{A} = \mathcal{M}_{t^\ast} \setminus {h}$, the available conditioning modalities.

\paragraph{Reconstruction Score.}
The held-out modality $h$ is reconstructed using all context modalities $\mathcal{A}$. Similarity $s_h$ is evaluated using either a normalized mean squared error, $\mathrm{NMSE}(I,\hat{I}) =
\left( \|I - \hat{I}\|_2^2 \right) / \left( \|I\|_2^2 \right)$, or multi-scale structural similarity (MS-SSIM). NMSE is used to account for the differing intensity ranges across modalities; a value close to zero indicates that the reconstruction error is small relative to the overall intensity magnitude of the ground truth. An MS-SSIM score close to one indicates high structural similarity between the reconstructed and ground truth modality at $t^*$.

\paragraph{Self-Consistency / Confidence Score.}
To estimate confidence, each context modality $m \in \mathcal{A}$ is in turn temporarily removed from the conditioning input and reconstructed from the remaining modalities $\mathcal{A} \setminus \{m\}$. For each modality, a similarity score $s_m$ is computed between reconstruction and ground truth. Similarity $s_m$ is evaluated using either a normalized reconstruction score MS-SSIM or NS, given by $\text{NS}(I,\hat{I}) = 1/(1 + \mathrm{NMSE}(I,\hat{I}))$. The value of NS being close to one indicates that the model reconstructs the held-out modality with similar quality to its reconstructions of the known modalities at $t^*$. The overall confidence at the hold-out time point is defined as the geometric mean of the similarity measures for the available sequences $\left(\prod_{m \in \mathcal{A}} s_m\right)^{1/|\mathcal{A}|}$.

\section{Experiments}
\subsection{Implementation Details}
The implicit neural representation is trained for 30,000 optimisation steps using the Adam optimiser with an initial learning rate of $10^{-3}$. The learning rate is reduced by a factor of two at 50\% and 75\% of training.For each patient, a separate model is trained on all but one timepoint, and each missing modality is reconstructed and evaluated at the held-out timepoint.

As a baseline, we employ linear interpolation (Interp) of the available modality intensities across time, evaluated at the held-out time point $t^\ast$.

\subsection{Results}
The INR outperforms the baseline across all modality and metric combinations (Tab.~\ref{tab:single_seq_mean_std}). The sole exception is for ADC evaluated with NS, where the interpolation baseline scores marginally higher, suggesting that the INR's gain over the interpolation baseline is smaller for ADC than for other modalities.

\begin{table}[h!]
\caption{Reconstruction performance (mean ± std.) for single-sequence holdouts, reported per method and modality.}
\centering
\small
\begin{tabular}{llllllll}
\hline
Method & Seq & n & NS$\uparrow$ & NMSE$\downarrow$ & MS-SSIM$\uparrow$ & RMSE$\downarrow$ & RMSE$_{\text{brain}}\downarrow$ \\
\hline

INR & T1CE  & 12 & \textbf{0.95 $\pm$ 0.02} & \textbf{0.06 $\pm$ 0.02} & \textbf{0.95 $\pm$ 0.02} & \textbf{0.02 $\pm$ 0.01} & \textbf{0.03 $\pm$ 0.01} \\
INR & FLAIR & 12 & \textbf{0.94 $\pm$ 0.03} & \textbf{0.06 $\pm$ 0.03} & \textbf{0.95 $\pm$ 0.03} & \textbf{0.03 $\pm$ 0.01} & \textbf{0.06 $\pm$ 0.04} \\
INR & ADC   & 12 & 0.84 $\pm$ 0.06 & \textbf{0.20 $\pm$ 0.09} & \textbf{0.82 $\pm$ 0.07} & \textbf{0.08 $\pm$ 0.03} & \textbf{0.15 $\pm$ 0.05} \\
INR & ALL   & 36 & \textbf{0.91 $\pm$ 0.06} & \textbf{0.11 $\pm$ 0.08} & \textbf{0.91 $\pm$ 0.08} & \textbf{0.04 $\pm$ 0.03} & \textbf{0.08 $\pm$ 0.06} \\
Interp & T1CE  & 9  & 0.81 $\pm$ 0.11 & 0.26 $\pm$ 0.18 & 0.85 $\pm$ 0.08 & 0.03 $\pm$ 0.01 & 0.05 $\pm$ 0.01 \\
Interp & FLAIR & 9  & 0.83 $\pm$ 0.08 & 0.21 $\pm$ 0.12 & 0.83 $\pm$ 0.13 & 0.04 $\pm$ 0.02 & 0.06 $\pm$ 0.03 \\
Interp & ADC   & 10 & \textbf{0.87 $\pm$ 0.05} & 0.16 $\pm$ 0.06 & 0.85 $\pm$ 0.06 & 0.07 $\pm$ 0.02 & 0.11 $\pm$ 0.02 \\
Interp & ALL   & 28 & 0.84 $\pm$ 0.08 & 0.21 $\pm$ 0.13 & 0.84 $\pm$ 0.09 & 0.05 $\pm$ 0.02 & 0.07 $\pm$ 0.03 \\

\hline
\end{tabular}
\label{tab:single_seq_mean_std}
\end{table}

We applied a Wilcoxon signed-rank test with Benjamini-Hochberg correction for multiple comparisons, using only matched patient--time point pairs with a minimum of three pairs per comparison. NS, NMSE, and MS-SSIM showed statistically significant differences between INR and Interp for T1CE and FLAIR ($p < 0.05$). No statistically significant differences were observed for ADC across any metric, nor for RMSE and RMSE$_{\text{brain}}$ for any sequence, the latter likely reflecting insufficient matched pairs rather than a true null effect.

Predicted confidence from the INR model showed strong association with true reconstruction quality across all held-out modalities, with high Pearson correlation for both NS ($r=0.933$--$0.996$) and MS-SSIM ($r=0.955$--$0.985$) (Tab. \ref{tab:confidence_agreement}). However, concordance analysis revealed modality-dependent calibration differences. For NS, agreement with the identity line was only moderate (CCC $=0.478$--$0.691$), indicating residual bias in error magnitude prediction despite near-perfect correlation for some sequences (e.g., FLAIR). In contrast, MS-SSIM confidence exhibited substantially stronger agreement (CCC $=0.680$--$0.933$), reaching excellent concordance for FLAIR and T1CE. These results suggest that the confidence estimator reliably ranks reconstruction quality across modalities, while absolute calibration is more accurate for MS-SSIM than NS.

\begin{table}[h!]
\centering
\caption{Agreement between predicted confidence and true reconstruction quality. Pearson correlation r measures association, while Lin’s concordance correlation coefficient (CCC) measures agreement with the identity line.}
\label{tab:confidence_agreement}
\setlength{\tabcolsep}{4.5pt} %
\begin{tabular}{l|cccc|cccc}
\hline
 & \multicolumn{4}{c}{\textbf{NS}} & \multicolumn{4}{c}{\textbf{MS-SSIM}} \\
\cline{2-5}\cline{6-9}
\textbf{Hold-out Seq} &
\textbf{Real} & \textbf{Pred} & \textbf{r} & \textbf{CCC} &
\textbf{Real} & \textbf{Pred} & \textbf{r} & \textbf{CCC} \\
\hline
ADC   & 0.800 & 0.838 & 0.933 & 0.691 & 0.759 & 0.903 & 0.955 & 0.680 \\
FLAIR & 0.719 & 0.905 & 0.996 & 0.478 & 0.834 & 0.891 & 0.984 & 0.905 \\
T1CE  & 0.894 & 0.826 & 0.993 & 0.527 & 0.884 & 0.879 & 0.985 & 0.933 \\
\hline
\end{tabular}
\end{table}

\begin{figure}[h]
\centering
\newcommand{\imw}{0.32\textwidth}

\begin{subfigure}{\imw}
\centering
\parbox{0.333\linewidth}{\centering\small GT}%
\parbox{0.333\linewidth}{\centering\small INR}%
\parbox{0.333\linewidth}{\centering\small Interp}\\[2pt]
{\setlength{\tabcolsep}{0pt}
\renewcommand{\arraystretch}{0}
\begin{tabular}{ccc}
\includegraphics[width=0.333\linewidth,height=0.333\linewidth]{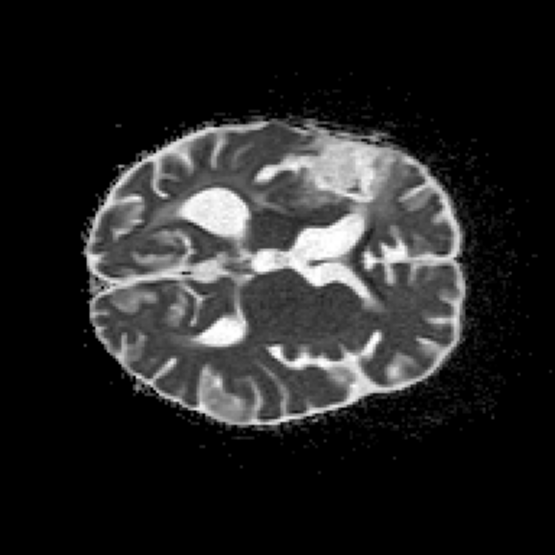}%
\includegraphics[width=0.333\linewidth,height=0.333\linewidth]{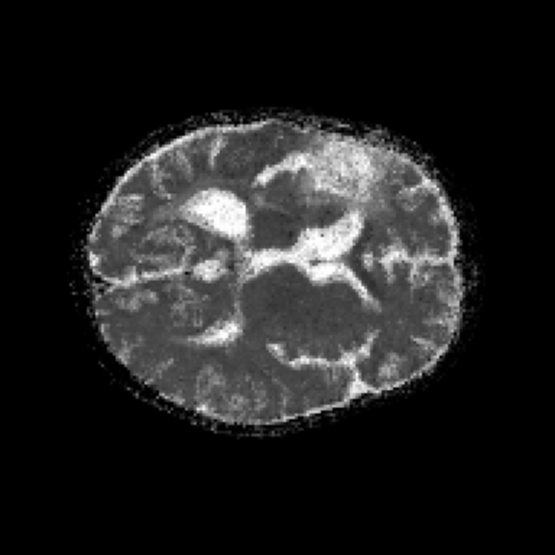}%
\includegraphics[width=0.333\linewidth,height=0.333\linewidth]{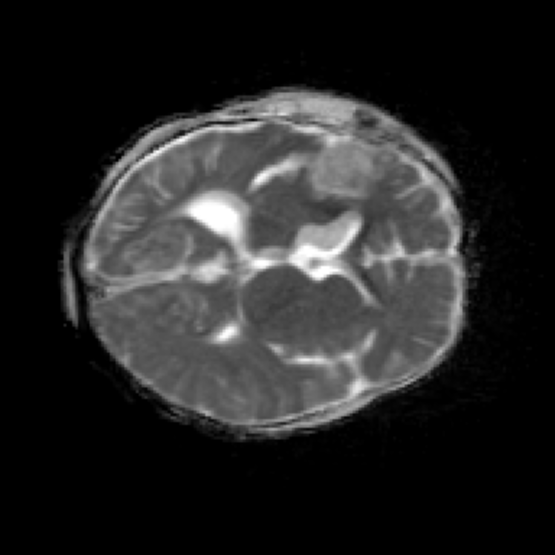}\\
\includegraphics[width=0.333\linewidth,height=0.333\linewidth]{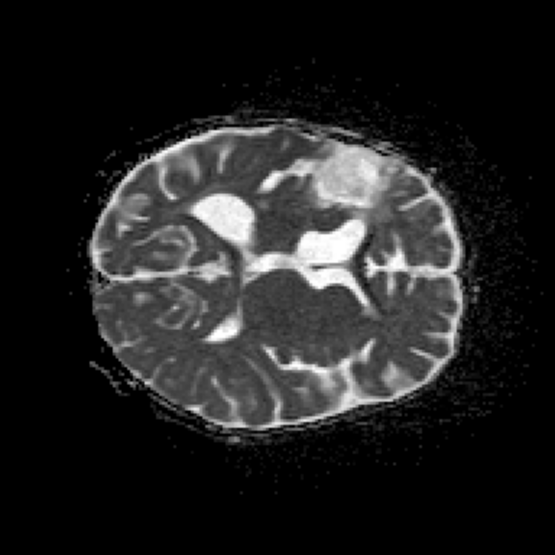}%
\includegraphics[width=0.333\linewidth,height=0.333\linewidth]{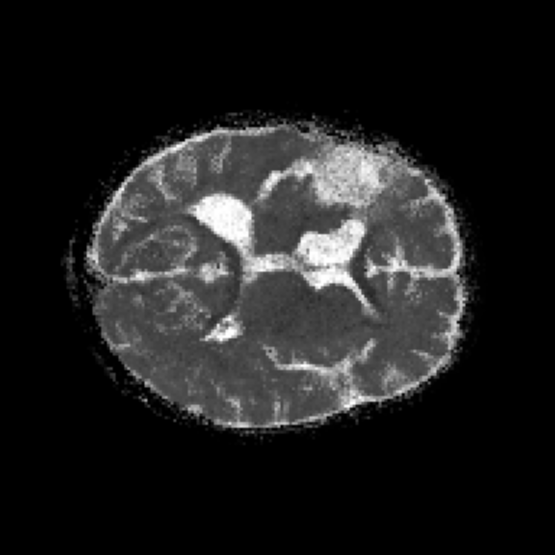}%
\includegraphics[width=0.333\linewidth,height=0.333\linewidth]{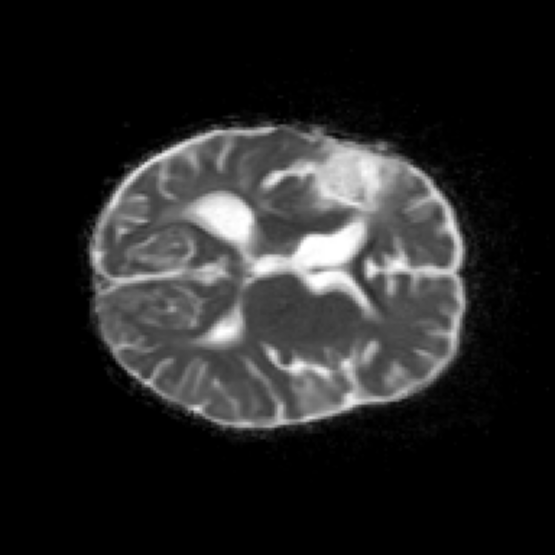}\\
\includegraphics[width=0.333\linewidth,height=0.333\linewidth]{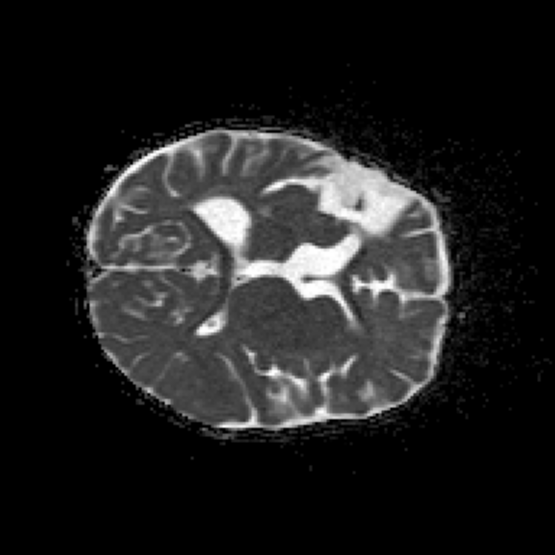}%
\includegraphics[width=0.333\linewidth,height=0.333\linewidth]{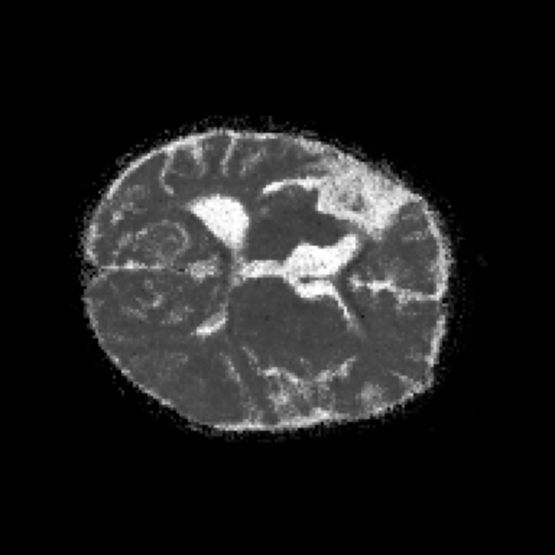}%
\includegraphics[width=0.333\linewidth,height=0.333\linewidth]{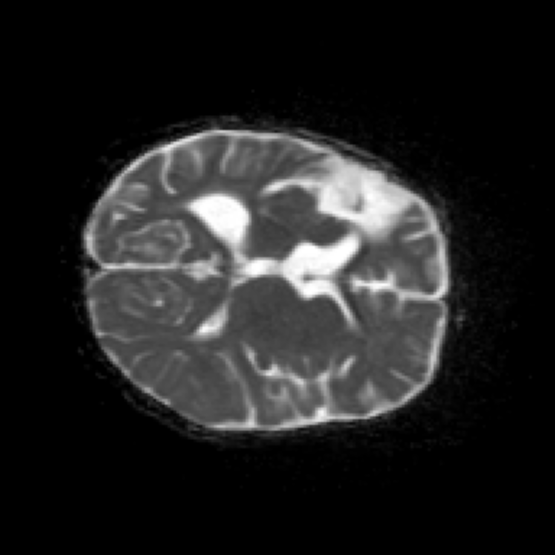}\\
\end{tabular}}
\caption{ADC}
\end{subfigure}\hfil
\begin{subfigure}{\imw}
\centering
\parbox{0.333\linewidth}{\centering\small GT}%
\parbox{0.333\linewidth}{\centering\small INR}%
\parbox{0.333\linewidth}{\centering\small Interp}\\[2pt]
{\setlength{\tabcolsep}{0pt}
\renewcommand{\arraystretch}{0}
\begin{tabular}{ccc}
\includegraphics[width=0.333\linewidth,height=0.333\linewidth]{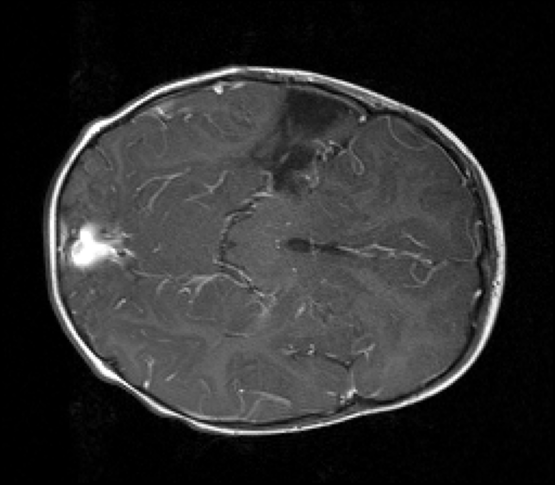}%
\includegraphics[width=0.333\linewidth,height=0.333\linewidth]{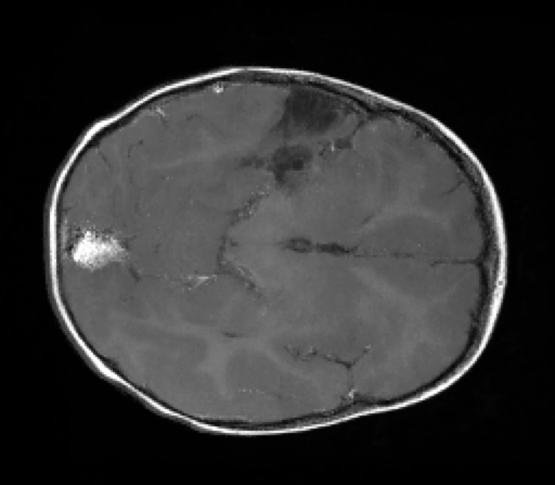}%
\includegraphics[width=0.333\linewidth,height=0.333\linewidth]{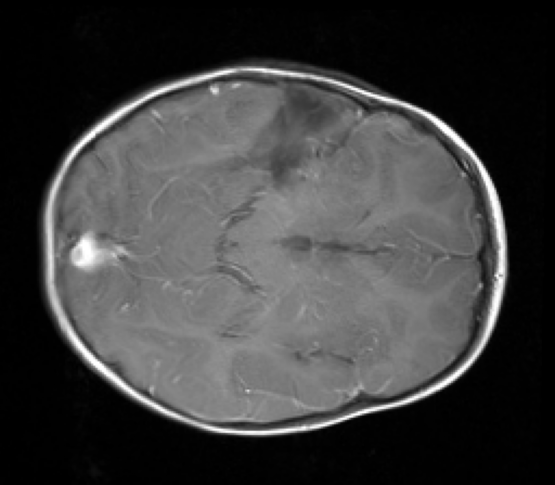}\\
\includegraphics[width=0.333\linewidth,height=0.333\linewidth]{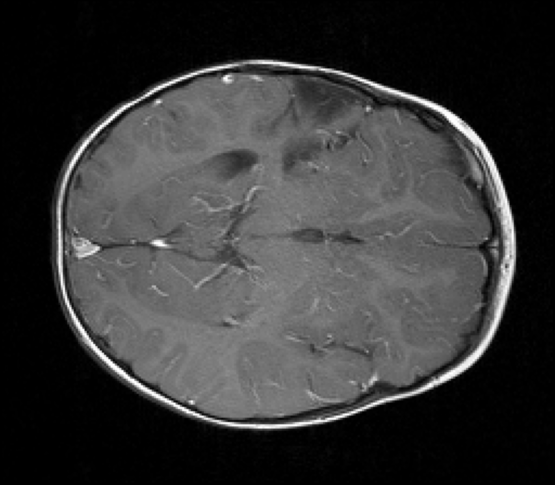}%
\includegraphics[width=0.333\linewidth,height=0.333\linewidth]{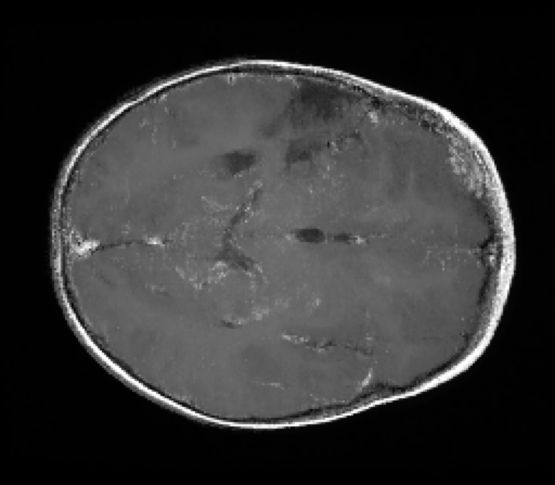}%
\includegraphics[width=0.333\linewidth,height=0.333\linewidth]{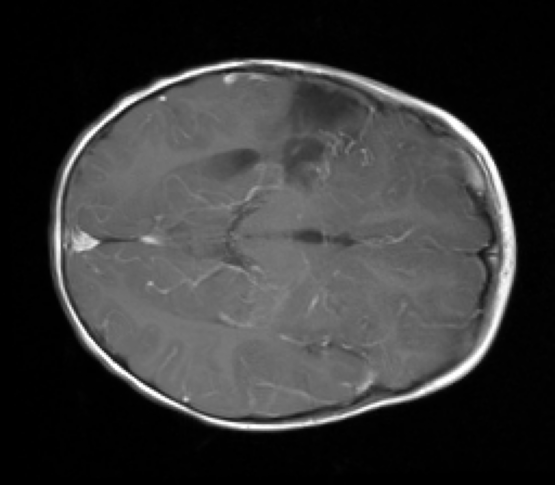}\\
\includegraphics[width=0.333\linewidth,height=0.333\linewidth]{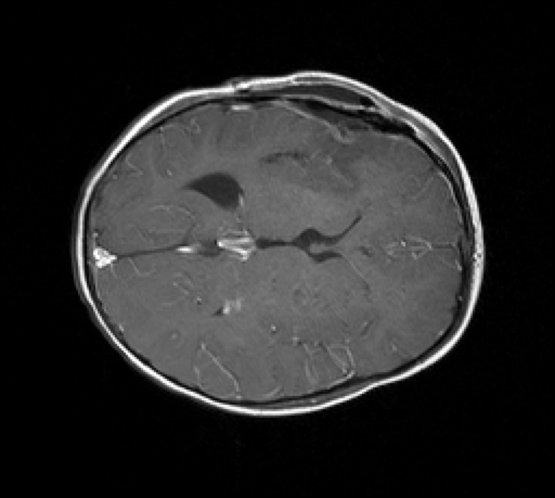}%
\includegraphics[width=0.333\linewidth,height=0.333\linewidth]{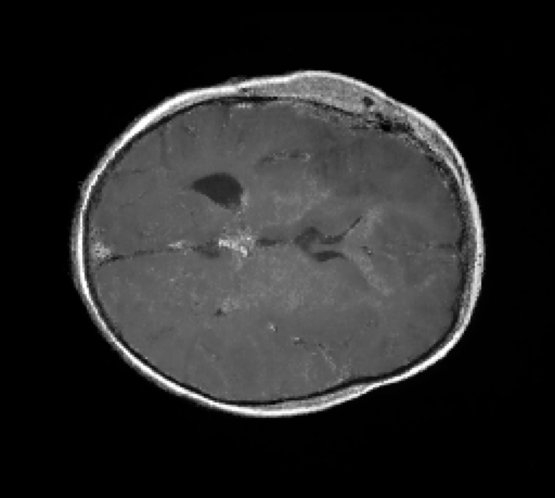}%
\includegraphics[width=0.333\linewidth,height=0.333\linewidth]{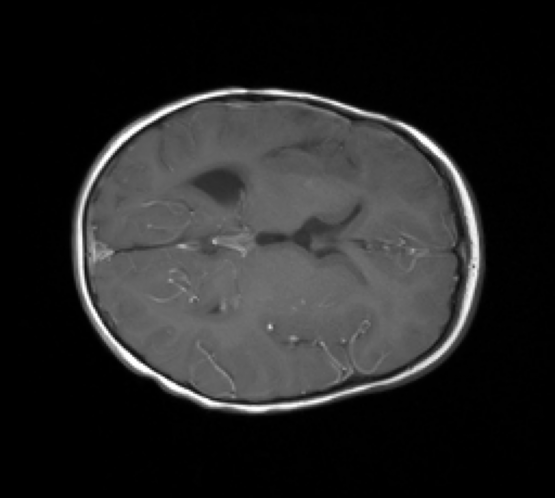}\\
\end{tabular}}
\caption{T1CE}
\end{subfigure}\hfil
\begin{subfigure}{\imw}
\centering
\parbox{0.333\linewidth}{\centering\small GT}%
\parbox{0.333\linewidth}{\centering\small INR}%
\parbox{0.333\linewidth}{\centering\small Interp}\\[2pt]
{\setlength{\tabcolsep}{0pt}
\renewcommand{\arraystretch}{0}
\begin{tabular}{ccc}
\includegraphics[width=0.333\linewidth,height=0.333\linewidth]{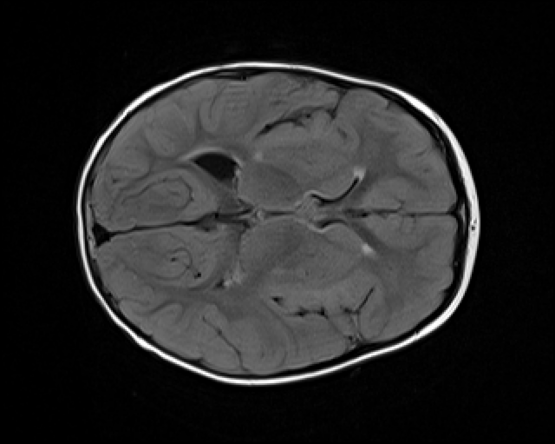}%
\includegraphics[width=0.333\linewidth,height=0.333\linewidth]{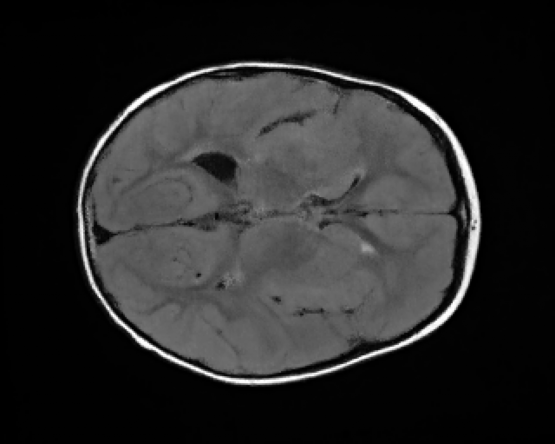}%
\vspace{-16pt}\hspace{1.5pt}\parbox[c][0.32\linewidth][c]{0.32\linewidth}{\vspace{-25pt}\centering\small N/A}\\
\includegraphics[width=0.333\linewidth,height=0.333\linewidth]{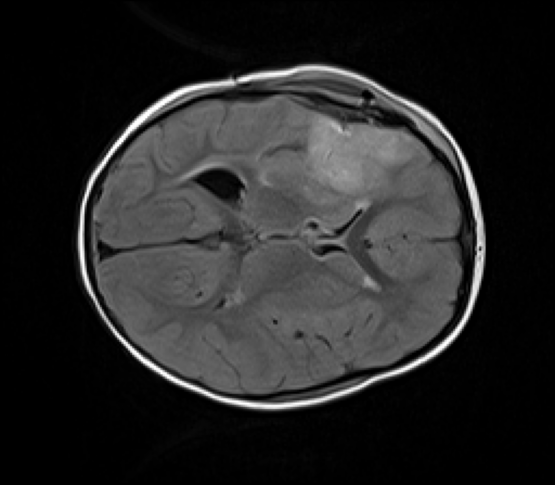}%
\includegraphics[width=0.333\linewidth,height=0.333\linewidth]{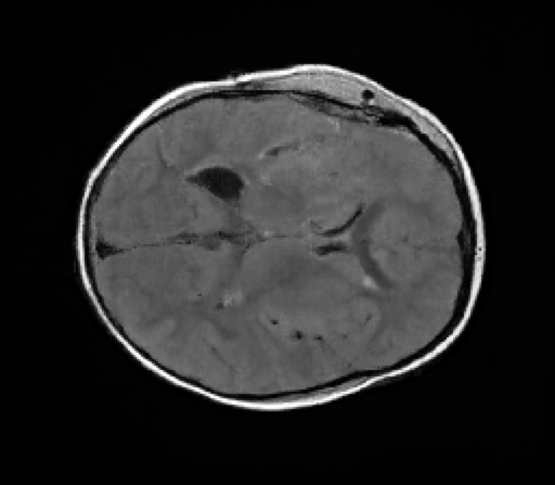}%
\includegraphics[width=0.333\linewidth,height=0.333\linewidth]{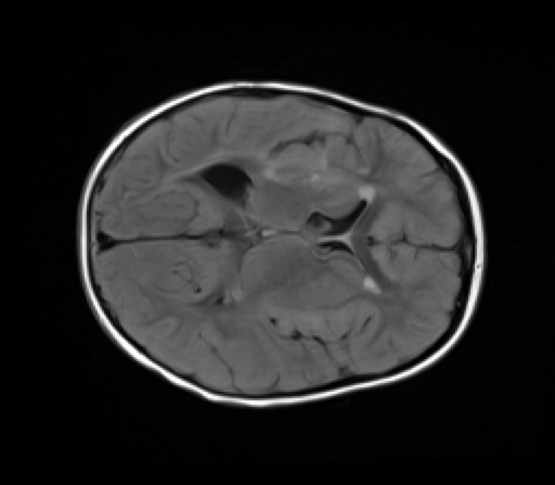}\\
\includegraphics[width=0.333\linewidth,height=0.333\linewidth]{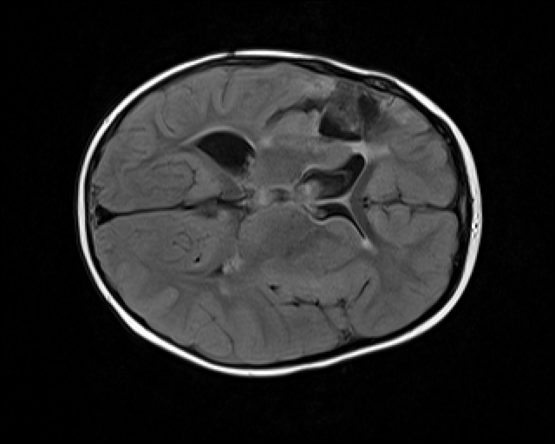}%
\includegraphics[width=0.333\linewidth,height=0.333\linewidth]{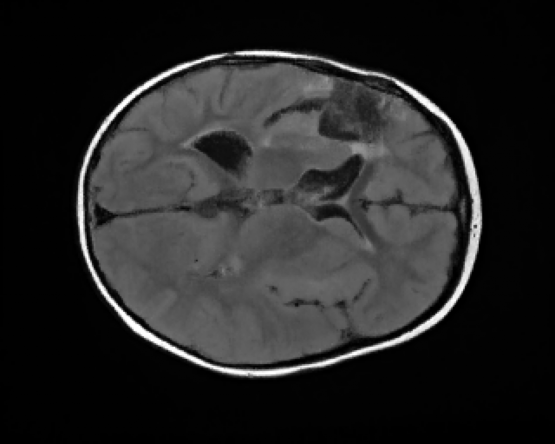}%
\includegraphics[width=0.333\linewidth,height=0.333\linewidth]{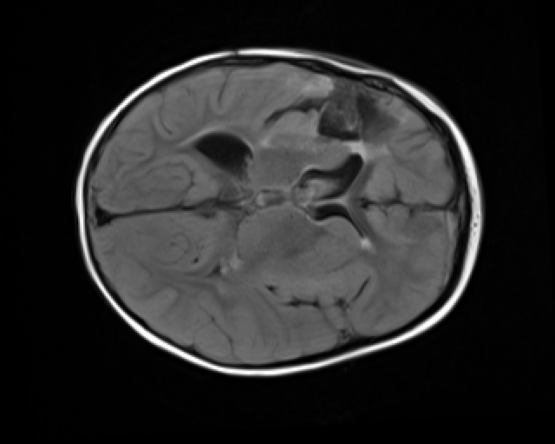}\\
\end{tabular}}
\caption{FLAIR}
\end{subfigure}

\caption{Qualitative comparison of GT, INR reconstruction, and linear interpolation 
baseline for one representative patient. Rows correspond to time points.}
\label{fig:qualitative}
\end{figure}

\section{Discussion}
\paragraph{Methodological Advantages}
The approach models multimodal longitudinal MRI as a continuous function of spatial world coordinates, time, and modality conditioning within a patient-specific INR. Unlike approaches that perform synthesis on discretised voxel grids~\cite{Chartsias,Liu}, the model operates in continuous coordinate space without resolution harmonisation. This avoids interpolation artefacts introduced by grid alignment and naturally accommodates heterogeneous spatial resolutions and acquisition protocols. The combination of multi-resolution hash encoding, Fourier temporal features, and stochastic modality dropout enables fine anatomical detail, longitudinal variation, and learning under incomplete modality configurations to be jointly modelled.

\paragraph{Limitations and Future Work.}
The current study evaluates the framework on a small patient cohort and focuses on interpolation at interior time points, limiting assessment of robustness to broader clinical variability and to temporal extrapolation beyond observed time points. Future work will investigate pretraining strategies, as well as longitudinal tissue modelling, towards full exploitation of longitudinal multimodal data for early recurrence detection in neuro-oncology.

\paragraph{Conclusion}
The results show that coordinate-based neural fields offer an effective representation for incomplete longitudinal MRI acquired under heterogeneous clinical conditions. The method achieves statistically significant improvements over linear interpolation for T1CE and FLAIR, with mean MS-SSIM of $0.95 \pm 0.02$ for T1CE compared to $0.85 \pm 0.08$ for the baseline. Reconstructions preserve anatomical detail without the blurring associated with voxel-wise linear interpolation, and the confidence measure correlates strongly with true reconstruction quality (Pearson $r$ up to $0.996$), suggesting potential for reliable deployment in heterogeneous clinical settings.

\begin{credits}
\subsubsection{\ackname} This research was partially funded by the Intramural Research Funding Grants ``AI-driven Longitudinal Lesion Tracking'' of the Faculty of Medicine, University of Augsburg, the Bavarian Center for Cancer Research as part of the Lighthouse ``Local Therapies'' and the Study Group ``Surrogate parameters for CNS Tumors in Childhood'', as well as by the Bavarian Ministry of Economic Affairs, Regional Development and Energy (StMWi) under grant number DIK-2310-0004//DIK0556/02.
\subsubsection{Disclosures.}
The authors used Claude Opus 4.8 (Anthropic) to assist with polishing the text of this manuscript. All AI-generated content was reviewed, verified, and edited by the authors.
\end{credits}

\end{document}